\documentclass[letterpaper, 10 pt, conference]{ieeeconf}
\usepackage{balance}
\usepackage{graphicx}

\usepackage{flushend}
\usepackage{cite}
\usepackage{multirow}
\usepackage{booktabs}
\usepackage{makecell}
\usepackage{verbatim}
\usepackage{amsmath}
\usepackage{ragged2e}
\usepackage{amssymb}

\IEEEoverridecommandlockouts                              

\title{\LARGE \bf
Repairing  Human  Trust  by  Promptly  Correcting  Robot  Mistakes  with An  Attention  Transfer  Model
}

\author{Ruijiao Luo$^{+}$, Chao Huang$^{+}$, Yuntao Peng, Boyi Song, Rui Liu$^{*}$

\thanks{Authors are with the Cognitive Robotics and AI Lab (CRAI), College of Aeronautics and Engineering,
        Kent State University, Kent, OH 44240, USA. $^{+}$ denotes that the first three authors have equal contributions to this paper. $^{*}$ Rui Liu is the corresponding author ruiliu.robotics@gmail.com }%
}

\begin{document}
\maketitle
\thispagestyle{empty}
\pagestyle{empty}

\begin{abstract}
In human-robot collaboration (HRC), human trust in the robot is the human expectation that a robot executes tasks with desired performance. A higher-level trust increases the willingness of a human operator to assign tasks, share plans, and reduce the interruption during robot executions, thereby facilitating human-robot integration both physically and mentally. However, due to real-world disturbances, robots inevitably make mistakes, decreasing human trust and further influencing collaboration. Trust is fragile and trust loss is triggered easily when robots show incapability of task executions, making the trust maintenance challenging. To maintain human trust, in this research, a trust repair framework is developed based on a human-to-robot attention transfer (\textbf{\textit{H2R-AT}}) model and a user trust study. The rationale of this framework is that a prompt mistake correction restores human trust. With \textbf{\textit{H2R-AT}}, a robot localizes human verbal concerns and makes prompt mistake corrections to avoid task failures in an early stage and to finally improve human trust. User trust study measures trust status before and after the behavior corrections to quantify the trust loss. Robot experiments were designed to cover four typical mistakes, \textit{wrong action, wrong region, wrong pose, }and \textit{ wrong spatial relation}, validated the accuracy of \textbf{\textit{H2R-AT}} in robot behavior corrections; a user trust study with $252$ participants was conducted, and the changes in trust levels before and after corrections were evaluated. The effectiveness of the human trust repairing was evaluated by the mistake correction accuracy and the trust improvement.
\end{abstract}

\section{Introduction}
With the integration of high-level human intelligence and low-level robot precision, human-robot collaborated systems (HRC) is widely used in various applications, such as daily elderly healthcare\cite{m1}, medical surgeries\cite{m2}, and advanced manufacturing\cite{m3}. A vital factor in ensuring a successful HRC is trust because an appropriate level of human trust largely facilitates the cooperation by avoiding unnecessary confirmations, interruption, and corrections. Trust reflects the willingness of dependency, a mutual understanding of intentions, and a positive prediction of the outcome \cite{b20,b5,b11,b21}. With a high-level trust, humans will closely collaborate with robots with high-level confidence and lower interruption frequency level. However, if a human over-trusts a robot, the human will not carefully monitor the robot's behaviors and approve inappropriate robot behaviors, causing severe consequences such as task failures and dangerous planning \cite{m4,m5}. If a human under-trusts a robot, the human will interfere frequently robot executions, interrupting the robot from doing its work \cite{b17,b19,confi}. Therefore, it is necessary to maintain an appropriate level of human trust for an effective HRC.

\begin{figure}[!t]
  \centering
 \includegraphics [width=0.92 \linewidth ]{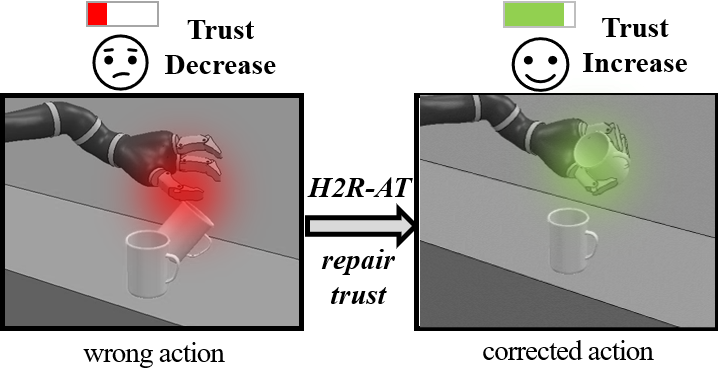}
  \caption{An illustration of the human trust repairing process by using the model \textbf{\textit{H2R-AT}}. The robot executions from the human perspective and the human trust level are presented as shown. By using the model \textbf{\textit{H2R-AT}} to correct robot execution errors, the human instructor's trust level in the robot is repaired. }
  \label{illu}
  \vspace{-0.6cm}
\end{figure}

However, maintaining an appropriate level of the human trust is challenging.
First, real-world factors, such as environmental disturbance and sensor uncertainties, inevitably cause robot mistakes \cite{m6,b27,b25,b26} and thus damage the human trust in robot capability. Frequent robot mistakes suggest unreliable robot executions and decrease human trust, impeding continuous collaboration. Second, abnormal and unexpected robot behaviors confuse humans about the robot's intent and performance, triggering unnecessary trust loss and human intervention. These abnormal and unexpected behaviors include slighted degraded action planning and obstacle-avoiding trajectory planning, which are necessary in tasks but may be misunderstood by a human. Frequent human interruptions increase robot computation and human cognitive load \cite{b18}, further reduce human trust.
Third, the trust level is dynamically adjusting and difficult to measure because the task types and difficulty, robot configurations, and human skill-level vary \cite{dd}.

To address the above challenges for trust maintenance, this paper developed a trust repair framework based on an attention-based behavior correction model - a human to robot attention transfer model (\textbf{\textit{H2R-AT}}) and a user trust study to repair trust when it is damaged by robot mistakes. As shown in figure \ref{illu}, with the model, the robot corrects its errors by interpreting the human verbal attention; the human trust will be repaired based on stably-good performance. This paper mainly has two contributions.

 \begin{figure*}[!t]
\centering
\includegraphics[scale = 0.4
]{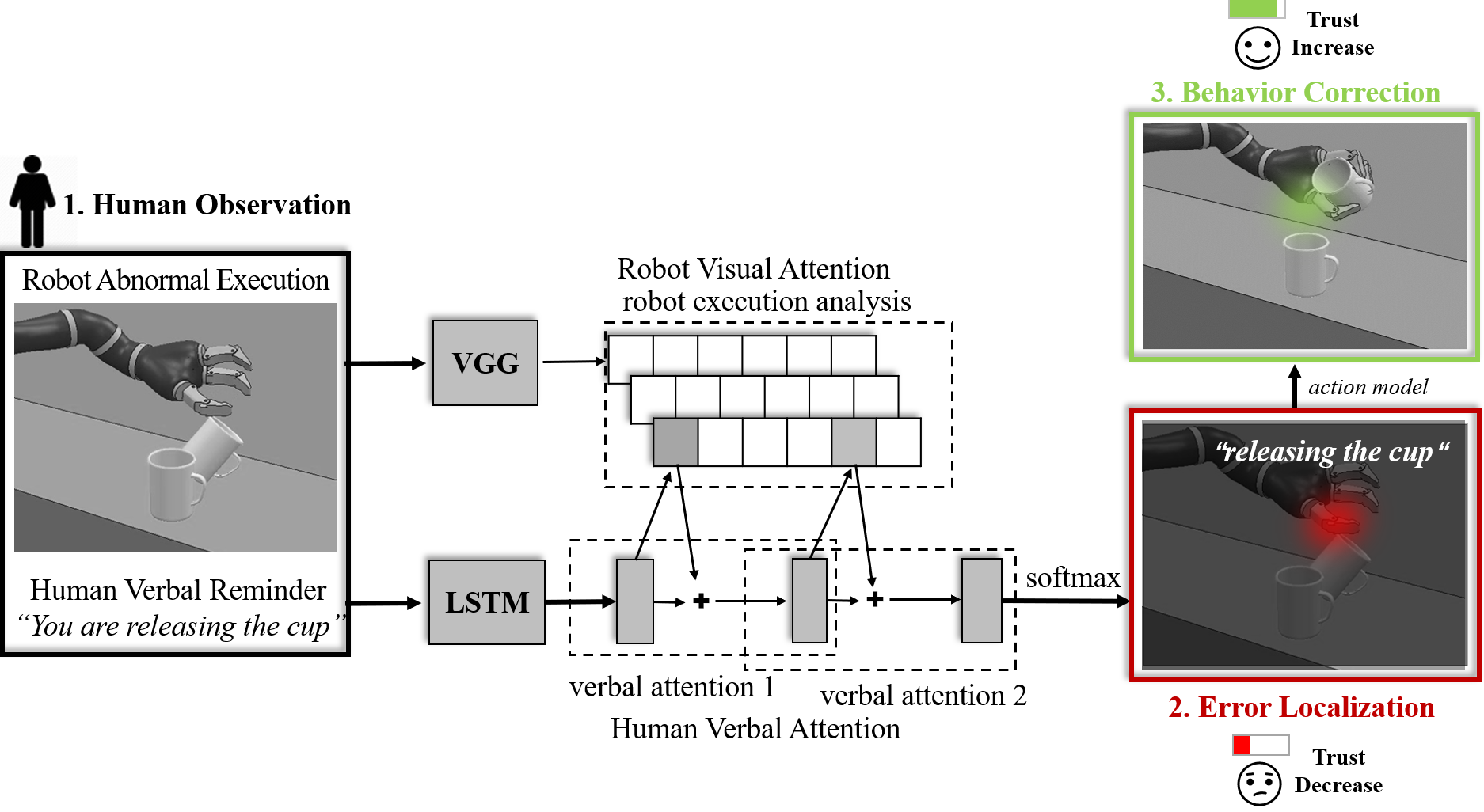}
\caption{ The framework of the methodology of using the \textbf{\textit{H2R-AT}} model to repair human trust by promptly correcting robot mistakes. First, at the notice of abnormal robot behaviors, the human instructor interrupts the robot execution and gives it an alert. Then, the \textbf{\textit{H2R-AT}} model combines feature vectors extracted from human verbal reminders and those from robot visual perceiving to get confined attention. With the attention, the robot can correct its abnormal behavior accordingly. Last, the human trust in the robot gets repaired as a result.}
\label{met}
\end{figure*}

\begin{itemize}
\item[$\bullet$] A novel framework of trust maintenance based on a novel attention transfer model, \textbf{\textit{H2R-AT}}, in human-robot interaction has been developed. Robot mistakes are correct timely; stably-good robot performance is maintained for a period; finally, human trust is repaired. Feasibility of the framework has been validated, providing theoretical support for a developing trustworthy interactive system.
\item[$\bullet$] A novel interactive trust-repairing user study has been developed with mixed measurements on both behavior and trust adjustment. This trust user study validated the feasibility of trust repairing after the behavior correction; it also makes a guideline for doing general trust repairing research in human-robot interaction, which is summarized as "identify mistakes, measure trust loss, correct behaviors, measure trust regain."
\end{itemize}

\begin{figure*}[!t]
\centering
\includegraphics [scale = 0.5
]{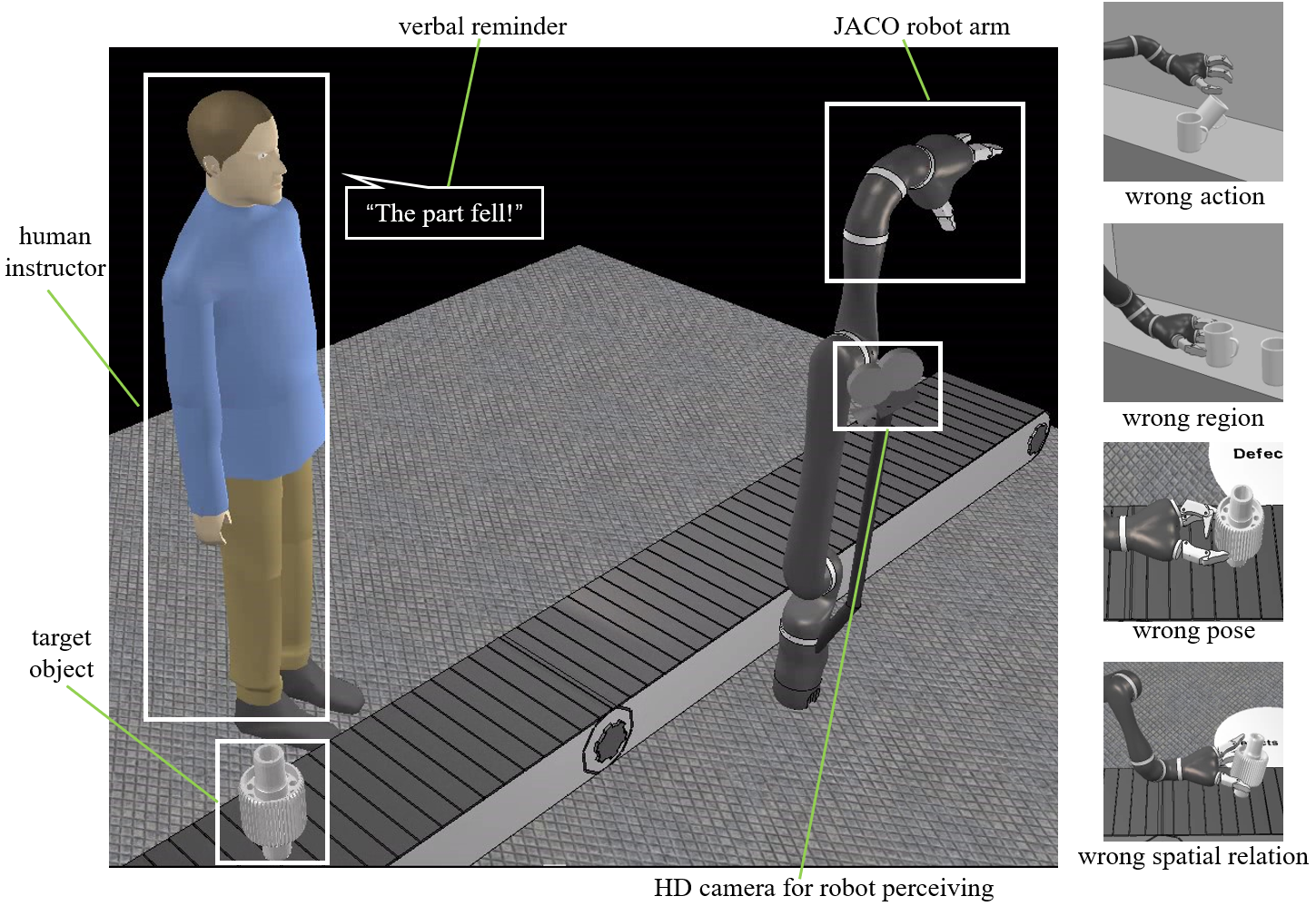}
\caption{The illustration of the experimental setup in the ``factory scenario", where the robot is supposed to pick up the target object. The human instructor gave a verbal reminder, and the HD camera recorded robot grasping.}
\label{whole}
\end{figure*}

\section{Related Work}
Recently attention has been investigated in HRC\cite{b40,b41}. The attention on hand gesture, gaze, and body language has guided an educational robot to focus and analyze student emotions, cognitive load, and the assistant request for recall and learning enhancement, improving the education effectiveness \cite{7723507}; the attention on human motion trajectories and head orientations guided a navigation robot to predict the future human trajectories, which is based on the result of the robots' planing its own trajectories and avoiding collision with the human \cite{b43,b44}. The attention on facial expression assisted a daily serving robot to adjusted the assistance timing and style, which increased the social acceptance of the robot\cite{b45,b46,b47}. Even though attention has been used in HRC research, a lot of the current work focuses on using attention for human instructions and plans understanding. In these processes, human is a cooperator who participates in task executions. As technology advances, there is a favorable trend of liberating human from interactions, at least from physical executions. It is more appealing that a human can only be a supervisor monitoring robot executions and alerting a robot only when its performance degrades and needs human assistance. Motivated by this, in this research, we explored another usage of attention in collaboration -using attention to express concerns on robot mistakes. In this research, our attention model is developed to both express concerns on current robot mistakes and localize the attention on where the mistake is. The ultimate goal is correcting robot mistakes and repairing human trust.

Trust modeling and the influential factors in HRC have been investigated recently \cite{lewis2018role}. It is challenging to observe and measure people's trust in robots since trust in HRC is multi-dimensional, incorporating both performance aspects (central in the human-automation literature) and moral aspects (central in the human-human trust literature) \cite{malle2020multi}. By collating 21 studies, \cite{hancock2011meta},  investigated the robot—related trust factors and found robot performance (such as reliability) had the strongest association with human trust; human-related factors (e.g., attitudes and comfort with robots) and environmental factors (e.g., culture and physical environment) contributed relatively less. \cite{broadbent2009acceptance} investigated trust factors in social situations such as elderly assistance, kid daily caring, and medical rehabilitation and found the human trust in robots was influenced by safety concerns and social acceptance. However, many of them focusing mainly on trust modeling, ignoring how to restore the human trust to ensure a continuous human robot interaction when it is damaged. Our trust repair framework considered the real-world scenarios where the robot mistakes that damage the human trust inevitably exists, while the trust repairing is critically important for continuing HRC.

\section{Trust Repairing by Correcting Errors}
Under the premise of human trust is decreased because of the abnormal robot behaviors as discussed above, the investigation of trust repairing process can be divided into two parts, the correction of abnormal robot behaviors and the trust level change after the correction. 
As shown in figure \ref{met}, when abnormal executions occur, the human instructor will immediately give a verbal alert to the robot. At the same time, the human instructor reports his current trust level in the robot. Then, the robot will interpret the human attention using the combination of the human verbal reminder and images it takes with the camera armed on it using the model \textbf{\textit{H2R-AT}}. After that, the robot will identify and correct the abnormal robot executions by localizing the human attention regions onto the actions it perceived. After seeing the robot corrects itself, the human instructor reports his current trust level in the robot again. The change in trust levels will be analyzed thoroughly with the given instructions to assess the model.

\subsection{Attention Supported Failure Avoidance}
Based on the stacked attention networks, the model \textbf{\textit{H2R-AT}} can extract the human attention and locate it on the specific region in robot perceiving. The human verbal reminders contain the message of the location and the type of robot behavior incorrectness. A Natural Language Process (NLP) module - long short-term memory (LSTM) is applied to interpret human reminders. In addition, the moment the human instructor raises the alert is considered to be the most suspicious moment to locate the corresponding human attention. Then, by using two layers of attention, \textbf{\textit{H2R-AT}} can transfer human attention underlying in the human reminders into the most critical region in the robot visual perceiving by gradually filtering out unrelated areas within its perceiving scope and finally focus on the abnormal regions. 

Next, to get a correction solution, a maximum likelihood recommendation method is used. The robot chooses the most recommended action $\hat{\alpha}$ according to $P(\alpha_i)$, the probabilities of each potential correction action $\alpha_i$. When the robot performs the chosen action and continues execution, the human can see that the robot listened to his instruction and corrected its errors.
\begin{equation}\label{eq:14}
    \hat{\alpha} = \mathop{\arg\max}_{\alpha}P(\alpha_i | \alpha_{attention}),
    i \in {1,2, ...} \\
\end{equation}

\subsection{Trust Measurement for Trust Repair}
In addition to logging human verbal reminders, our study also queried the human’s trust assessments. This factor is used to evaluate the effectiveness of \textbf{\textit{H2R-AT}} in human trust repairing. In this paper, we use a self-reported measure of trust, which is similar to \cite{Xu2015op} and asked the human instructors to report their current trust level towards the robot before and after the error correction. 

These trust assessments are queried using a modified Visual Analog Scale (VAS) \cite{Reips2008i}, as shown in Fig. 4. The human trust level is divided into five scales, \textit{completely distrust, distrust, neutral, trust}, and \textit{completely trust} for the human instructors to choose from. The trust level of \textit{completely trust} means both the success of the execution and the absence of abnormal executions. In order to better show the changes in human trust levels before and after the error correction. The results are generalized to continuous curves by using mixed-Gaussian functions. As shown in the equation \ref{level}, when \textit{i} is equal to 1, 2, 3, 4, and 5 respectively, it corresponds to \textit{completely distrust, distrust, neutral, trust}, and \textit{completely trust}; the variance of each Gaussian Function is fixed to 1; $x$ represents the numerical score for different trust assessments, for example, $x_1 = 1, ..., x_5 = 5$; \textit{N$_i$} denotes the number of participants who chose \textit{i} and \textit{$N$} denotes the total number of participants; $N(x)$ represents the percentage of participants whose trust assessment score is equal to $x$.

\begin{equation}\label{level}
N(x) = \sum_{i=1}^5 \frac{N_i}{N}\cdot\frac{exp\{-\frac{(x-x_i)^2}{2}\}}{\sqrt{2\pi}},
i \in {1, 2, 3, 4, 5}
\end{equation}

\section{Experiment Settings}
The effectiveness of repairing human trust via \textbf{\textit{H2R-AT}} model was evaluated by the effectiveness of robot failure avoidance and the reliability of raising human trust.

\subsection{Robot Behavior Simulation}
To learn and validate the effectiveness of the \textbf{\textit{H2R-AT}} model in guiding human-robot trust repair, two robot task scenarios, "serve water for a human in a kitchen" and "pick up a defective gear in a factory", were applied. 
As shown in figure \ref{whole}, four representative abnormal robot execution cases were simulated. Participants assessed their trust before and after the correction. Task scenarios were designed with the CRAIhri simulation framework, which is developed based on V-REP simulation software, JACO robot arm model, and machine learning interface. In the experiment, a human instructor was involved in monitoring the robot and gave it an alert when recognizing abnormal behaviors. The robot visual perceiving when the human sends the alert was recorded for analysis. By using the \textbf{\textit{H2R-AT}} model to align both robot visual perceiving and human verbal alerts, the nonlinear relation between human attention and robot attention was modeled to guide robot failure avoidance. Decision-making strategies were delivered to the robot by establishing real-time communication between V-REP and Python. 

\subsection{Human User Study} 
A user study was conducted on the crowd-sourcing platform Amazon Mechanical Turk \textit{d4}. In total, $252$ English-speaking participants were recruited with $\$1.5$ payment each.  About $12,000$ verbal reminders were collected to label $12,000$ most-typical images of abnormal robot executions. Results and evaluations will be detailed in the next section. 

The participants were asked to watch the videos of robot performing the assigned tasks and give instructions in natural language when they noticed an abnormal behavior to help the robot correct it. During each task in the survey, the participants were asked to assess their trust levels in the robots twice, before and after the robot correct its abnormal behaviors. 


The user study consisted of two parts, a tutorial as the  experiment instruction and an actual survey for the data collecting. The tutorial includes two cases, successful and unsuccessful grasping. Given a robot performance, the participants were suggested to choose \textit{"Completely Distrust"} or \textit{"Distrust"} from \textit{"Completely Distrust", "Distrust", "Neutral", "Trust", and "Completely Trust"}.In the actual survey for data collecting, the participants were required to monitor the robot executions through $10$-second videos. When the participant noticed an abnormal robot behavior happening, he was supposed to alert the robot in natural language to help it correct its error and then complete its task successfully. They also reported the region where they focused on when the abnormal behavior occurs and their trust levels towards the robot before and after correction. The actual survey had four tasks, covering four basic robot error types, \textit{wrong action, wrong region, wrong pose, and wrong spatial relation}, and two simulated scenarios, \textit{the kitchen scenario and the factory scenario}. Multiple factors related to the response behaviors of the volunteers, including the accuracy record on the MTurk, the time duration for answering questions, the empty answer rate, irrelevant question percentages, etc are considered to control the quality of the survey data.

\section{Result}




\subsection{Accuracy of Failure Avoidance}

Unlike human attention, which concentrates on one area, robot attention is distributed in several regions due to the model and data uncertainty. 
Results show that 
73.68\% of human verbal attention was correctly transferred to a robot as human expected; Besides, the probability of the most chosen solution, which would be adopted for the robot to correct the abnormal behavior, was $90.75\%$. Then, the accuracy of robot failure avoidance was $66.86\%$, showing the largely improved robot performance under human guidance. 
This result shows the potential of using \textbf{\textit{H2R-AT}} model to avoid robot failure in the real-world environment. 

\begin{figure}[!t]
\centering
\includegraphics [
scale = 0.6
]{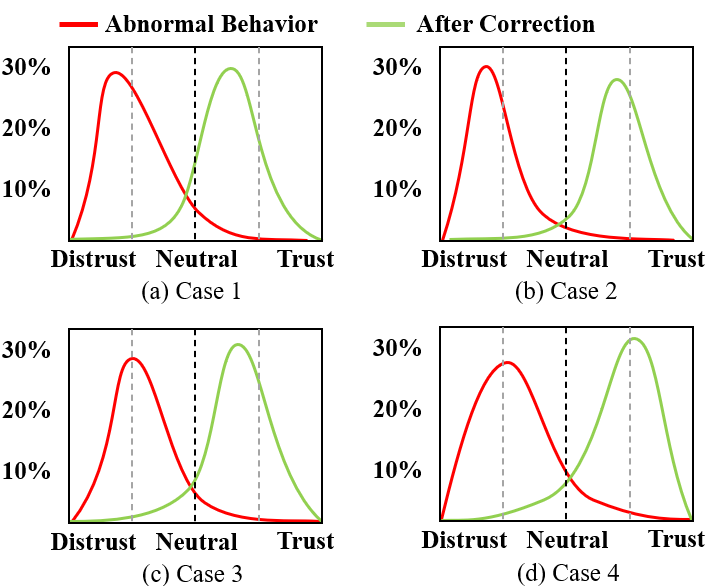}
\caption{The portion of the people with different trust levels. The model \textbf{\textit{H2R-AT}} successfully raised the human trust level from \textit{distrust} (shown in the red line) to \textit{trust} (shown in the green line). Case 1: Wrong action; Case 2: Wrong region; Case 3: Wrong pose; Case 4: Wrong spatial relation,
}
\label{1to4}
\end{figure}

\subsection{Trust Repairing through Error Reduction}
As mentioned in the "Trust Measurement for Trust Repairer Evaluation" section, the human trust levels were quantified and generalized for investigation. The distribution of the reported trust levels before and after correction for each case is shown in Figure \ref{1to4}. Overall, the human trust level has shown a successful increase compared to that before error correction. Yet, the trust levels were not fully restored. The detailed result analysis is shown below.

\begin{table}[]
\centering
\caption{\scshape Mann-Whitney U Test for Trust Levels}

\vspace{6mm}
\centering
\begin{tabular}{cp{0.17\textwidth}p{0.07\textwidth}c}
\toprule
\multirow{2}{*}{\textbf{Case}} &
 \multicolumn{2}{c}{\textbf{Median Trust Level}} & \multirow{2}{*}{\textbf{P}}\\ 
 & Before & After & \\
\midrule
\textbf{1}                     & Completely Distrust & Trust & \multirow{4}{*}{\textless .001} \\
\textbf{2}                     & Completely Distrust & Trust &                                 \\
\textbf{3}                    & Distrust            & Trust &                                 \\
\textbf{4}                     & Distrust            & Trust &  \\  
\bottomrule
\multicolumn{4}{c}{(a) Trust Levels Before and After Correction}\\
\end{tabular}

\vspace{4mm}
\centering
\begin{tabular}{p{0.11\textwidth}p{0.11\textwidth}p{0.11\textwidth}}
\toprule

\multicolumn{2}{c}{\textbf{Case (Average Trust Level)}}         & \makecell[c]{\textbf{P}}                  \\
\midrule
\makecell[c]{\textbf{1} (1.336)}               & \makecell[c]{\textbf{2} (1.252)}              & \makecell[c]{0.0936}             \\
\makecell[c]{\textbf{3} (1.510)}               & \makecell[c]{\textbf{4} (1.600)}              & \makecell[c]{0.1824}             \\
\makecell[c]{\textbf{1} and \textbf{2} (1.294)}         & \makecell[c]{\textbf{3} and \textbf{4} (1.555)}        & \makecell[c]{\textless .001}     \\
\bottomrule
\multicolumn{3}{c}{\centering{(b) Average Trust Levels Before Correction}}\\
\end{tabular}

\vspace{4mm}
\centering
\begin{tabular}{p{0.11\textwidth}p{0.11\textwidth}p{0.11\textwidth}}
\toprule

\multicolumn{2}{c}{\textbf{Case (Average Trust Level)}}         & \makecell[c]{\textbf{P}}                  \\
\midrule
\makecell[c]{\textbf{1} (4.174)}               & \makecell[c]{\textbf{2} (4.148)}              & \makecell[c]{0.4614}             \\
\makecell[c]{\textbf{3} (4.187)}               & \makecell[c]{\textbf{4} (4.194)}              & \makecell[c]{0.3550}             \\
\makecell[c]{\textbf{1} and \textbf{2} (4.161)}         & \makecell[c]{\textbf{3} and \textbf{4} (4.191)}        & \makecell[c]{0.2451}     \\
\bottomrule
\multicolumn{3}{c}{\centering{(c) Average Trust Levels After Correction}}\\
\end{tabular}

\label{utest}

\vspace{2 mm}
\justifying{This table presents the results of the Mann-Whitney U test for trust levels for result analyze. Case 1, 2, 3, and 4 sequentially denote the 4 cases the participants were showed, where case 1 and 2 took place in the kitchen, while case 3 and 4 in the factory. Table (b) presents the Mann-Whitney U test for trust levels before error correction. Table (c) presents the Mann-Whitney U test for trust levels after error correction. Thus, the first row in Table (b) and Table (c) means the comparison between the two cases in the kitchen scenario, the second row in the factory scenario, and the last row means the comparison between the kitchen scenario and the factory scenario.}
\end{table}

Mann-Whitney U test was applied in result analysis to see the progress in repairing trust. The result shows that for all four cases, participants trusted the robot significantly more after the abnormal execution was eliminated (p\textless.001), shown in table \ref{utest} (a). It proves that in this experiment, human trust levels were raised in all cases, which illustrates the significant efficacy of \textbf{\textit{H2R-AT}}. However, compared to the preset human trust level, the average repaired trust level after robot behavior correction dropped. The trust level was restored to $73\%$ of its initial level. The quality of the trust repairing result shows consistency. As shown in table \ref{utest} (c), on the scale one to five, the average trust level for each corrected case was $4.174$, $4.148$, $4.187$, and $4.194$, respectively. As shown in table \ref{utest} (c), the Mann-Whitney U test result states there was no distinct difference in the trust level for corrected executions between different cases. It means that within each scenario, after correction, human trust was restored to a similar level. It means that as long as the errors were eliminated, participants were likely to hold a higher level of trust towards the robot. This is intuitive since all four corrected executions showed successful outcomes without any dangerous signs that could weaken human trust.

\section{Conclusion}
In this paper, the human trust repairing mechanism was analyzed through robot error reduction using the attention transfer model \textbf{\textit{H2R-AT}} based on stacked neural networks. 
This mechanism was evaluated by an experiment in which participants were recruited to monitor the robot executions before and after behavior correction and assess their trust level at that moment. 
All participants reported a higher level of trust after the correction, which shows the success of human trust repair. Several determining factors for trust level change were also given with exemplification in this work.

\end{document}